# Semantic Instance Labeling Leveraging Hierarchical Segmentation

Steven Hickson
Georgia Institute of Technology
shickson@gatech.edu

Irfan Essa
Georgia Institute of Technology
irfan@cc.gatech.edu

Henrik Christensen
Georgia Institute of Technology
hic@cc.gatech.edu

## Abstract

*Most of the approaches for indoor RGBD semantic labeling focus on using pixels or superpixels to train a classifier. In this paper, we implement a higher level segmentation using a hierarchy of superpixels to obtain a better segmentation for training our classifier. By focusing on meaningful segments that conform more directly to objects, regardless of size, we train a random forest of decision trees as a classifier using simple features such as the 3D size, LAB color histogram, width, height, and shape as specified by a histogram of surface normals. We test our method on the NYU V2 depth dataset, a challenging dataset of cluttered indoor environments. Our experiments using the NYU V2 depth dataset show that our method achieves state of the art results on both a general semantic labeling introduced by the dataset (floor, structure, furniture, and objects) and a more object specific semantic labeling. We show that training a classifier on a segmentation from a hierarchy of super pixels yields better results than training directly on super pixels, patches, or pixels as in previous work.*

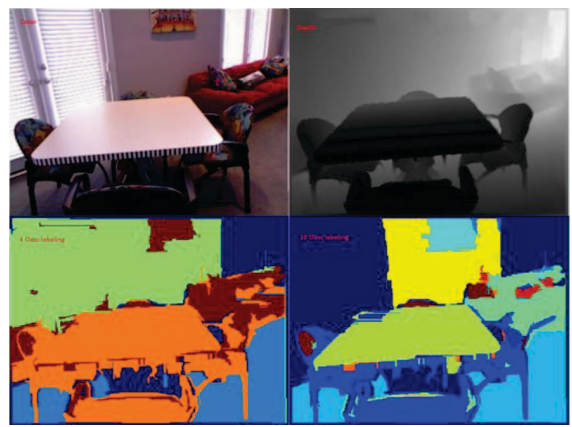

Figure 1. Semantic segmentation and labeling of 3D Point clouds. Top: RGB and depth maps used as input in the algorithm. Bottom: The 4 Class output and 10 Class output as described in Section 5.

## 1. Introduction

Recently, much work has been done on dense semantic labeling and indoor scene understanding for the use of robots. However, these are often limited to scene recognition, limited classes, or structural information. With the introduction of affordable RGBD cameras such as the Microsoft Kinect, dense point clouds can be constructed in indoor environments with minimal effort. This kind of data has changed the way we do object detection and labeling. Although object recognition has come a long way with these new sensors, indoor semantic labeling and scene understanding is still in it's infancy. For robots to effectively interact with an indoor environment autonomously, they need access to accurate semantic information about their environment. Some work has been done with labeling of SLAM maps [16, 18, 24, 12, 22, 15]. However, most previous work doesn't focus on dense semantic maps to allow a robot to interact with the environment.

With the introduction of the NYU Dataset [17], we now have an indoor densely labeled Kinect dataset for scene understanding. Our implementation aims to tackle the problem of accurate, dense, and fast semantic segmentation. Our method uses hierarchical segmentation to construct meaningful features from full objects and not patches or individual pixels. The assumption being that whole objects will have more meaningful features (shape, size, color) than superpixels, patches, or pixels that correspond to parts of objects. We evaluate our method using the NYU Dataset [17] and segment each frame as either *floor, structure, furniture, or props*. We additionally segment each frame using more specific classes to label objects such as *bed, chair, etc.* based on [5].

Our method is novel in that we focus on instance labeling based on a hierarchy of segmented super pixels. This has many advantages to previous work using bounding boxes or pixel-based detectors. We aim to not only label each segment as belonging to a category but to separate each segment as different objects even if they belong to the same category. In a scene, we can not only know what pixels are labeled as furniture, we can know how many pieces of furniture there are, where they are, and what pixels belong to each one. Because of this, we avoid the pitfalls of bounding-box detectors in that we know the contour of the object and we avoid the pitfalls of general semantic segmentation since we generate instances of objects. Another advantage arises due to this; since we test on larger, merged segments from super pixels, our testing time can be very fast.

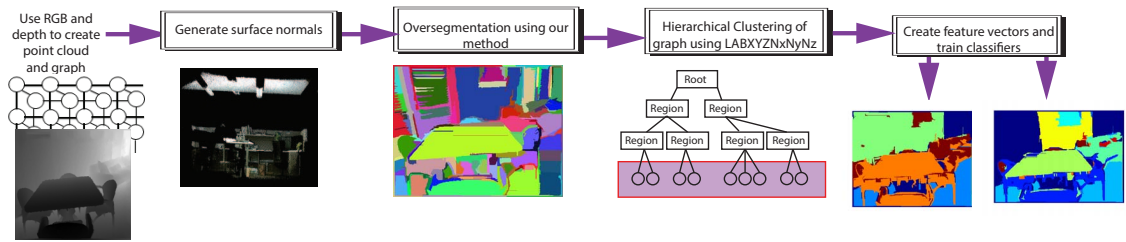

Figure 2. An overall schematic of our full method. First we generate the point cloud and a graph, then we generate surface normals, then we over segment the point cloud, create a hierarchy of super pixels, and finally train our feature vectors using a random forest of decision trees.

## 2. Related Work

Most previous work has been focused on semantically labeling each pixel. Recently this has been extended to be each super pixel or patch or label using convolutional neural networks or conditional random fields. In the past, this was mostly done in outdoor datasets and on RGB images instead of 3D data. On 3D indoor scenes, most of the work has been focused on categorizing the environment and that work which does semantically label uses pixel or super pixels to train classifiers.

In [16], the authors focus on labeling scenes only (not objects or segments) such as office, kitchen, hallway, etc; [18] and [22] label scenes the same way with an actual robot. Nuchter et al. [18] only label objects as the wall, floor, ceiling, or doors. These are classes that don't fit the challenging categorization we are trying to achieve. [12] generates meaningful and accurate semantic labeling; however, it trains specific objects (such as a bed or printer) and tests them assuming they are in the same environment. So although they can determine useful objects in an environment, they require knowledge of the current room. Richtsfeld et al. [21] use NURBS fitted patches to find graspable objects but do not classify them.

Silberman et al. [17] introduced the NYU dataset and a method to semantically label indoor scenes. They use a contour based method with structural inference and a set of various supervised learning features to train a SVM. Gupta et al. [8, 9] discuss using bounding box detectors for instance labeling and using an improvement of [17] with geometric encoding and an additive SVM kernel for semantic labeling.

Couprie et al. [5] use a multi-scale convolutional neural network to learn unsupervised features and impose labels on super pixels. They also introduced a new subset of the NYU V2 dataset [17] to test, which is based on the labels occurring most frequently in the dataset. This allows us to determine how an algorithm does labeling more specific categories. Wang et al. [28] also use an unsupervised method but instead use multi-modality learning in order to learn features that a linear SVM is trained on to generate semantically-labeled super pixels. Lin et al. [13] focus on recognizing cuboid objects by extending CPMC [4] to leverage 3D data and then use a conditional random field to generate semantic labels. Cadena et al. [3] similarly use a conditional random field on superpixels in order to train general semantic labels. Stuckler et al. [26] use temporal information to create an object-class segmentation using SLAM and a random decision forest in order to generate a fully labeled 3D map.

## 3. Method

Our method is the first to use a hierarchy of super pixels to train a classifier on the NYU V2 dataset. We rely more on the shape, color, surface, and position of objects than hand-crafted features like SIFT or pixel-based metrics like conditional random fields or convolutional neural networks. The theory behind this is that people learn objects based on these characteristics not patches of objects. This methodology requires a higher level segmentation than just super pixels. We use a variant of the hierarchical segmentation method based on Hickson et al. [10] in order to create segments that match objects instead of patches. The full method is shown in Figure 2.

## 3.1. Fast Surface Normal Estimation

As Kinect data is extremely dense, we use surface normals to help segment the 3D data and train our classifier. We use the surface normal estimation described in [11]. It uses integral images to estimate dense surface normals at every point for projective devices and is implemented in PCL [23]. Surface normals allow us to enhance the segmentation of [10] and still maintain the non-linear combination of color and depth. Our enhancement of the method is described in Section 3.2.

## 3.2. 3D/4D Segmentation

Our segmentation method is based on the method described in [10] that uses a hierarchical graph-based segmentation based on [7]. Although there are other methods that segment 3D data such as [25, 29, 1, 29, 27, 19] that could have been used, we decided on [10] based off it's impressive results and open source library.

Hickson et al. [10] build a graph in which each node corresponds to a *toxel*, which they define as a temporal voxel. In that method, in each graph, each node is connected to it's 26-neighbors in two different ways. The first edge's weight is the absolute difference in the depth of the two nodes: $|D(x,y,z,t) - D(x',y',z',t')|$, where $D(x,y,z,t)$ is the depth value in the spatiotemporal volume at $time = t$ with the corresponding $x$, $y$, and $z$ location. The neighborhoods are defined as $(x',y',z',t') \in \mathcal{N}(x,y,z,t)$. The regions are then merged according to [7].

In [10], after the algorithm produces a segmentation using only the depth, another graph is created with the same nodes as before using the second set of edges. The second edge weight is defined as the difference in the color of the two nodes: $|C(x,y,z,t) - C(x',y',z',t')|$, where $C(x,y,z,t)$ is the LAB color value in the spatiotemporal volume at $time = t$ with the corresponding $x$, $y$, and $z$ location. As before, edges are created for each of the 26-neighbors in the same neighborhood space, creating an over-segmentation of super-toxels.

Hickson et al. [10] show that merging depth and color separately is better than combining them in a linear manner. We modify their method to allow a single, non-linear edge weight. Our method differs from [10] in that we construct a graph using only 8-neighbors (ignoring time and using only voxels) whilst maintaining non-linearity. We use the surface normals estimated in Section 3.1 and the color to create one edge weight $W$:

$$W = max(\sqrt{N(x,y,z)^2 - N(x',y',z')^2}, \sqrt{C(x,y,z)^2 - C(x',y',z')^2}) \quad (1)$$

where $N(x,y,z)$ is a function that yields the surface normals from the point cloud, $C(x,y,z)$ is a function that yields the colors from the point cloud and $(x',y',z') \in \mathcal{N}(x,y,z)$, where $\mathcal{N}(x,y,z)$ is the neighborhood of voxel $(x,y,z)$. We then over-segment the same way as in [10].

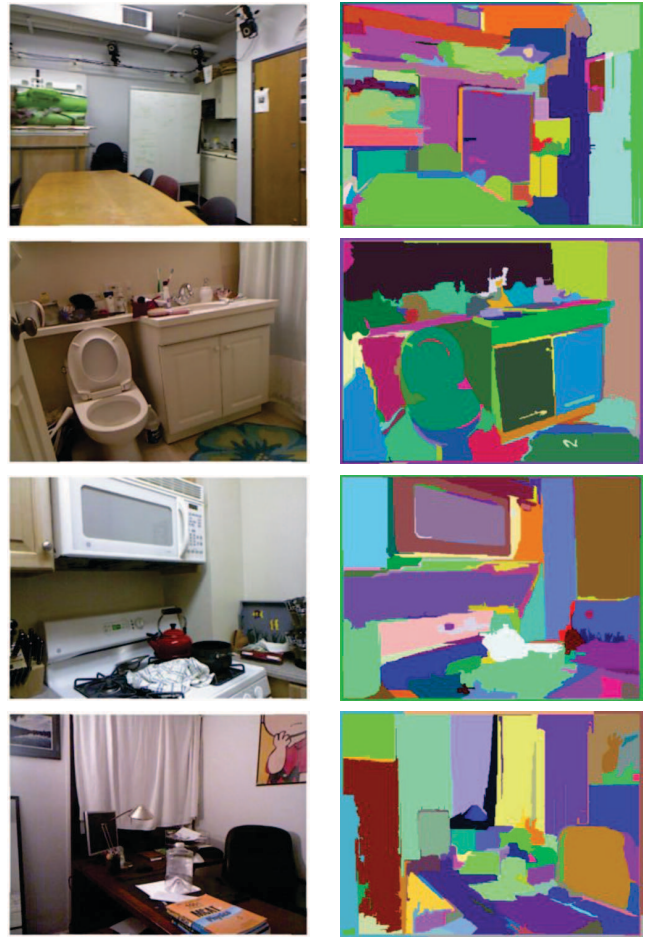

Figure 3. Hierarchical segmentation results on the NYU Dataset [17], Left: rgb, Right: our method's pseudocolored segments.

## 3.3. Hierarchical Construction

After the over-segmentation described in Section /ref-sec:seg, we compute feature vectors for each super-voxel that we use for hierarchical propogation and for training our semantic classifier. For feature vectors we extend the feature space of [10] by using histograms of CIE Lab color, histograms of 3D position, and histograms of 3D surface normals, called LABXYZNxNyNz histograms, using all the voxels in each region. As in [10], rather than computing large multi-dimensional histograms, we compute 1-dimensional histograms for each feature. It is important to note that we have added surface normals as features here and are ignoring optical flow as we are just looking at single RGBD frames. We use 20 bins of LAB color features and 30 bins of the XYZ position features just like in [10], adding 30 bins of the NxNyNz surface normal features. We also compute the region's 3D size, height, width, 2D centroid, 3D centroid, 3D minimum, and 3D maximum as extra features.

Using the aforementioned feature vectors, a new graph ([10] calls it the S-graph) is created to propagate the hierarchy. The graph is comprised of vertices that are each super-voxel region and edges that are formed to join neighboring regions using Equation 2, which has been modified from [?] to use surface normals instead of optical flow. We then construct a dendrogram that is a hierarchical tree of super-voxel regions defined by similarity where the root node is the entire-set merged and the leafs are the super-voxels computed by Section 3.2.

To match the regions we use a modified version of the SAD difference equation in [10] that adds surface normals instead of optical flow. The difference between Region $R$ and Region $S$ is defined as:

$$\Delta H = \sum_{i=1}^{\text{NUMBINS}} |\frac{R_l[i]}{R_N} - \frac{S_l[i]}{S_N}| + |\frac{R_a[i]}{R_N} - \frac{S_a[i]}{S_N}| \\ + |\frac{R_b[i]}{R_N} - \frac{S_b[i]}{S_N}| + |\frac{R_x[i]}{R_N} - \frac{S_x[i]}{S_N}| + |\frac{R_y[i]}{R_N} - \frac{S_y[i]}{S_N}| \\ + |\frac{R_z[i]}{R_N} - \frac{S_z[i]}{S_N}| + |\frac{R_nx[i]}{R_N} - \frac{S_nx[i]}{S_N}| + |\frac{R_ny[i]}{R_N} - \frac{S_ny[i]}{S_N}| \\ + |\frac{R_nz[i]}{R_N} - \frac{S_nz[i]}{S_N}| \quad (2)$$

where $R_N$ is the number of voxels in Region $R$ and $S_N$ is the number of voxels in Region $S$.

We use the same parameters defined in [10] with a lower tree cut of 0.15. Pseudo-colored output of this high level segmentation on the NYU V2 dataset is shown in Figure 3.

### 3.4. Feature Selection

Feature selection for this task is extremely complex and it is difficult to determine which properties might be important. The objects in each category can vary quite a bit and it is difficult to tell what features are best for this classification. Some semantic segmentation methods [20, 17, 8] also use custom, expert features to label classes using the NYU dataset [17] while others use machine learning techniques to learn features such as Couprie et al.[5], which uses a convolutional neural network to learn important features. The arguments between the benefits and limitations of these approaches are beyond the scope of this paper. For each region, we use general features such as shape, size, position, orientation, surface normals, and color in order to catch any important features. In addition we test the features described in Section 3.3 as well as the impact of adding a 1000 cluster bag-of-words of SIFT feature points.

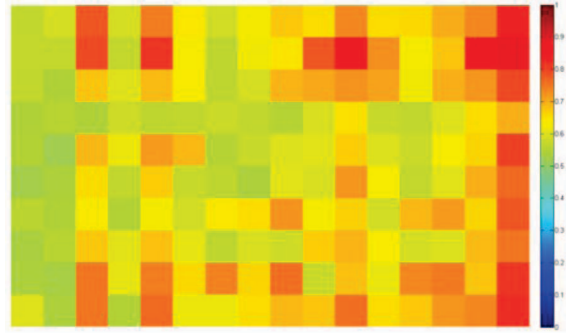

Figure 4. Classification Accuracy for different 3D Features on a validation set as described in Section 3.4. Heat mapped with blue being 0% and red being 100%

A random forest of 500 decision trees yielded the best results on our test using the NYU Depth Dataset V2 [17]. When testing all the different features, a random forest was the most accurate classifier. This makes some amount of sense given that we are using histograms and the features are very independent of each other. This was the reasoning behind using a random forest as opposed to an SVM or neural network.

For the histogram features described in Section 3.3, we ran all different classifiers available in Matlab on each different feature as a binary classifier using a random validation set from the training data of the NYU dataset [17]. The results are shown in Figure 4 with 1 being 100% accuracy and 0 being 0% accuracy for the top 10 classes where each column is a feature from a subset of our features (size, 2D X centroid, 2D Y centroid, 3D X centroid, 3D Y centroid, 3D Z centroid, A histogram, B histogram, L histogram, Nx histogram, Ny Histogram, Nz histogram, 3D X histogram, 3D Y histogram, 3D Z histogram, and all combined features) and each row is a class. The last column is a combination of all the features and is shown to do the best. Not only did we find that the combination of all features yields the best result, we also found that random forests [2] always yield the best accuracies for our data. This confirms our hypothesis that random forests will work best in this scenario.

## 4. Experiments

We tested our method on two different tasks. The first is semantically labeling everything as either *floor, structure, furniture, or props* as proposed in [17]. This is shown in Section 4.1 and reveals how our method generalizes categories accurately compared to state of the art. The second test is labeling the 13 most frequent classes, with the 14th being other, in the NYU V2 dataset as proposed by [5]. This is shown in Section 4.2 and reveals how our method does on more specific objects and classifications.

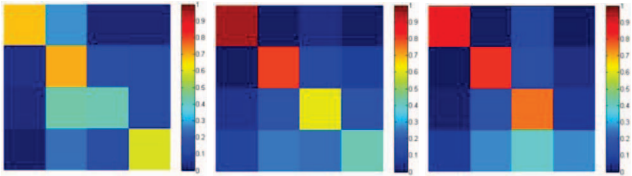

Figure 5. Confusion Matrix for 4 classes of NYU Dataset Left: [17], Center: our method with SIFT, Right: our method without SIFT. Heat mapped with blue being 0% and red being 100%

### 4.1. 4 Class tests

After determining our features and classifier, we ran our method on the NYU V2 dataset [17] and trained the 4 Class and 14 Class classifiers to compare against previous implementations .We further experiment by training our method on the 4 Class set with and without SIFT [14]. Previous methods such as [17, 20, 8] use a BOW of SIFT features [6] to help classify. Being uncertain about the usefulness of SIFT in such a general classification method, we trained our method with and without it as a test. The class pixel accuracy confidence matrices for our method (with and without a BOW of SIFT) and NYU's method [17] are shown in Figure 5. It is around 10% more accurate although we suffer categorizing props, showing no improvement over the 42% of [17]. In table 1, our method is compared to other state of the art methods showing us to have better accuracy in ground, furniture, overall class accuracy, and overall pixel accuracy. Note that we do not compare to [8] in this table since they use a different accuracy metric then all of the other papers testing on this dataset.

### 4.2. 14 Class tests

We also test our method using the 14 classes (including other) proposed by Couprie et al [5]. It is important to note that this classifier was trained independently of the 4 class classifier does not use the classifications shown in Section 4.1 although that may improve classification results. For this test, we did use a BOW of SIFT features as we are finding specific objects. We compare here against Wang et al. [28] and Couprie et al. [5], which are both unsupervised learning methods as opposed to out method which uses hand-crafted features. As shown in Table 2, we improve total class accuracy by more than 5% over Wang et al. [28] and over 11% more than Couprie et al. [5]. We do slightly worse in a couple categories such as bed, sofa, wall, books, and TV. It is uncertain whether our performance increases are due to the hand-crafted features or the hierarchical segmentation. However, since Silberman et al. [17] use a much larger set of features that include some of ours, and both methods outperform [17], we suspect that most of our accuracy gain comes from using a hierarchical segmentation.

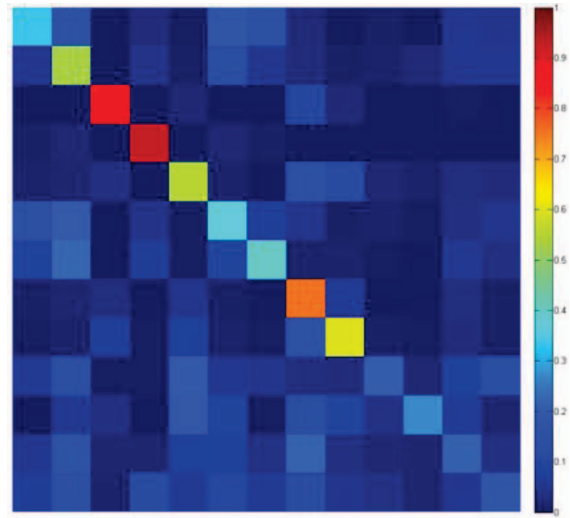

Figure 6. Confusion Matrix for the top 13 classes of NYU Dataset[17] as proposed by [5] (ignoring the 14th class other). Heat mapped with blue being 0% and red being 100%

## 5. Results

To compliment the quantitative results shown in Table 1 and Table 2, we show some qualitative results of our algorithm in Figures 7 and 8. Figure 7 shows the result of our algorithm on the 4 general classes and Figure 8 shows the result of our algorithm on the 14 specific classes.

For both tests, we excel at labeling different structures, the floor, the ceiling, and different furniture. However, our algorithm suffers labeling objects. This could be due to the hierarchical segmentation. It is possible that small objects get merged to larger segments, which would label them as part of the furniture or structure near the object. It is also possible that our generic features involving shape and surface normals do not generalize well for objects. The lack of textural information could also be contributing to the lack of object accuracy.

However, as shown in Table 2, the hierarchical segmentation does allow the algorithm to excel at recognizing the furniture and the structure. Our method is far superior at recognizing chairs, the ceiling, the floor, tables, and windows. These all have relatively similar shapes, sizes, and surface normals. Graph-based segmentation methods also tend to match the boundary of more rectangular objects, creating better segments for planar objects, which contributes to our high accuracy in these areas.

## 6. Conclusions/Future Work

In this paper, we present a new approach to indoor semantic labeling that uses a hierarchical segmentation method with general features pertaining to shape, size, and

1072

|  | Ground | Furniture | Props | Structure | Class Acc. | Pixel Acc. |
|---|---|---|---|---|---|---|
| Silberman et al. [17] | 68 | 70 | 42 | 59 | 59.6 | 58.6 |
| Couprie et al. [5] | 87.3 | 45.3 | 35.5 | **86.1** | 63.5 | 64.5 |
| Cadena et al. [3] | 87.9 | 64.1 | 31 | 77.8 | 65.2 | 66.9 |
| Stuckler et al. [26] | 90.7 | 68.1 | 19.8 | 81.4 | 65.0 | 68.1 |
| Wang et al. [28] | 90.1 | 46.3 | **43.3** | 81.4 | 65.3 | N/A |
| Our Method (No SIFT) | **95.3** | 60.9 | 42 | 80.2 | **69.6** | 69.5 |
| Our Method | 88.5 | **75.5** | 27.1 | 81.8 | **68.2** | 71.8 |

Table 1. A comparison of the per-pixel and per-class classification accuracy of the 4 class set comparing our algorithm to other state of the art methods.

|  | Couprie et al. [5] | Wang et al. [28] | Our Method |
|---|---|---|---|
| bed | 38.1 | **47.6** | 33.5 |
| chair | 34.1 | 23.5 | **53.3** |
| ceiling | 62.6 | 68.1 | **84.8** |
| floor | 87.3 | 84.1 | **92.8** |
| picture | 40.4 | 26.4 | **55.3** |
| sofa | 24.6 | **39.1** | 36.8 |
| table | 10.2 | 35.4 | **40.1** |
| wall | **86.1** | 65.9 | 75.6 |
| window | 15.9 | 52.2 | **59.8** |
| books | 13.7 | **45.0** | 20.4 |
| TV | 6.0 | **32.4** | 27.3 |
| Furniture | **42.4** | 16.7 | 21.5 |
| Objects | 8.7 | 12.4 | **17.6** |
| Class Acc | 36.2 | 42.2 | **47.6** |

Table 2. A comparison of the per-pixel and per-class classification accuracy of the 14 class set proposed by [5] comparing our algorithm to other state of the art methods.

color in order to generate meaningful segments. We then use a random forest of decision trees in order to train a classifier to classify both general categories and more specific categories on the NYU V2 dataset. We show that our results are comparable and even improve on state of the art methods including those that use convolutional neural networks, super pixels, and advanced structural features. Future work could include GPU optimization, imposing structural constraints as in [17, 8], and combining the labels with Kinect Fusion or another SLAM system to make a completely labeled model of the environment.

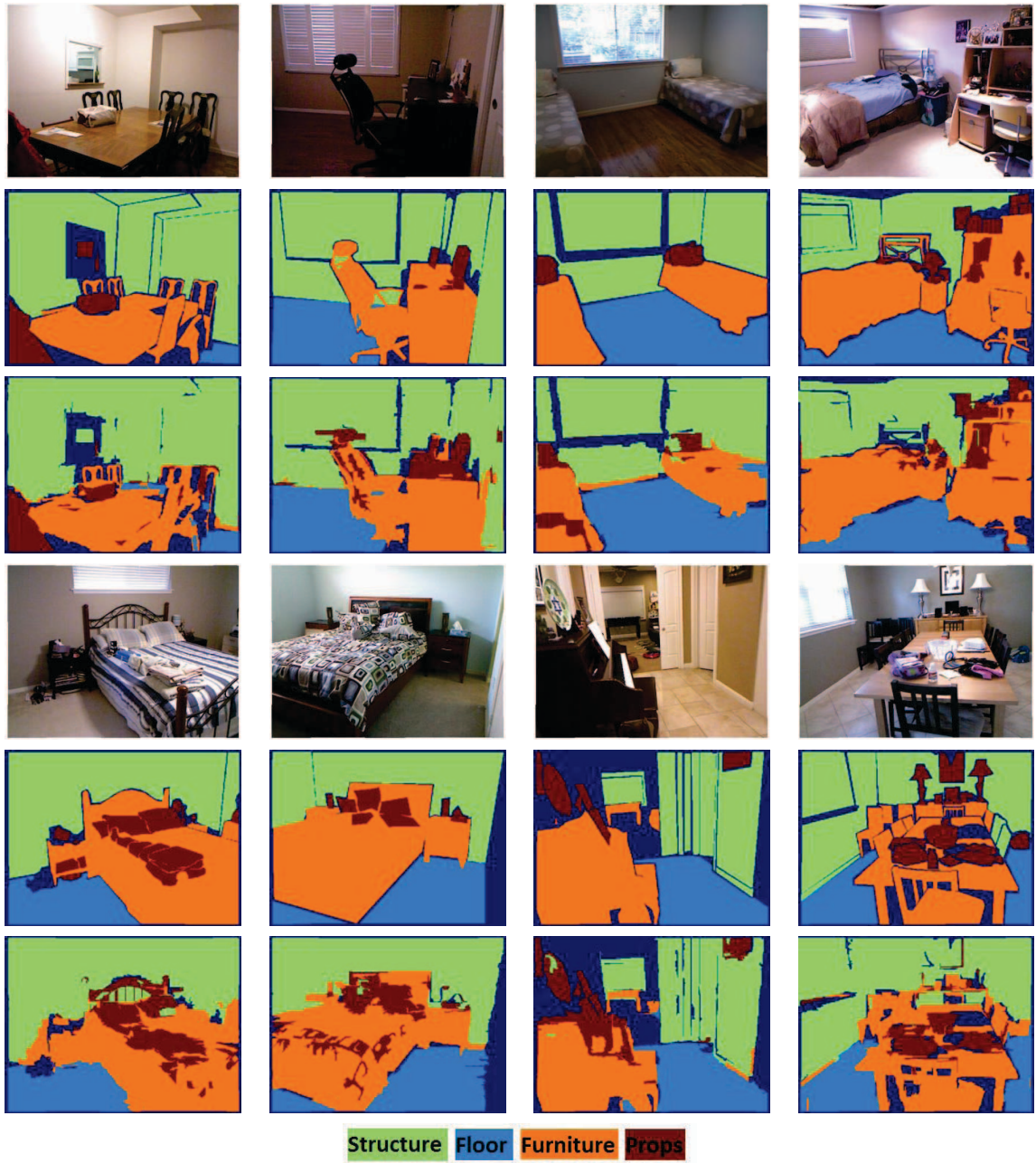

Figure 7. Results from the 4 Class set from [17]. The top row shows RGB, the middle row shows the ground truth labels, and the bottom row shows our classification.

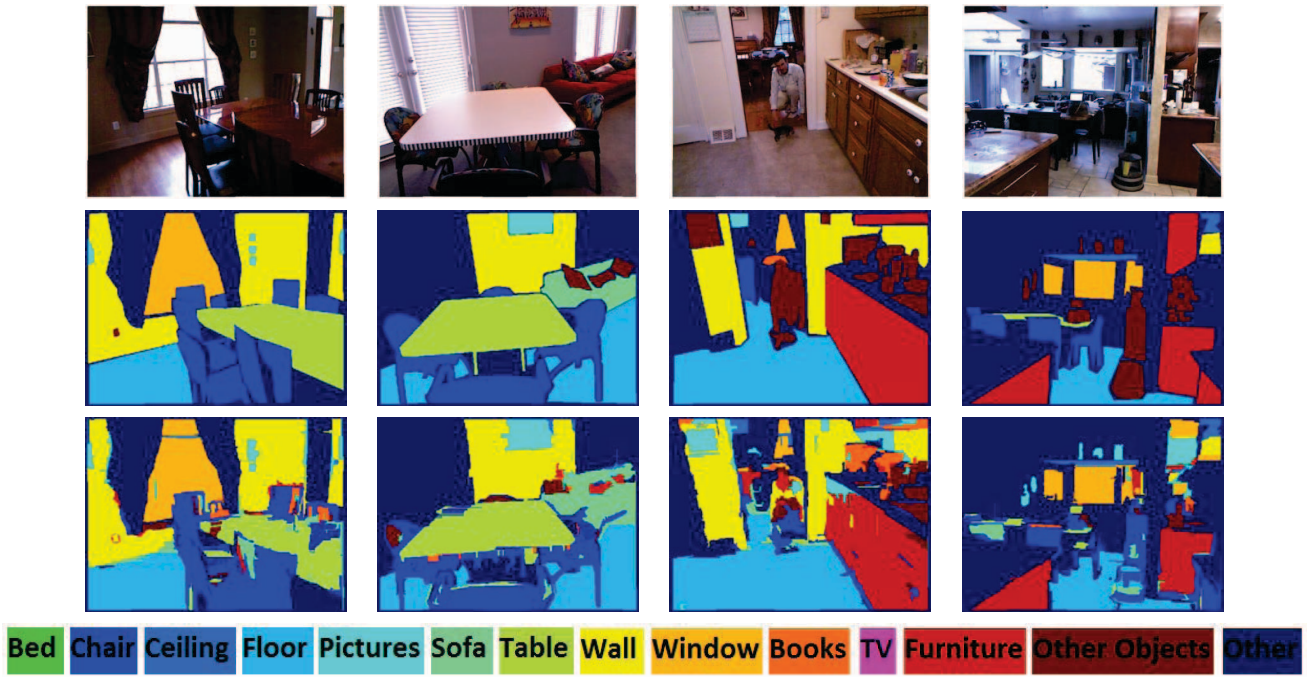

Figure 8. Results for the 14 class set proposed by [5] The top row shows RGB, the middle row shows the ground truth labels, and the bottom row shows our classification.